\documentclass[letterpaper]{article} % DO NOT CHANGE THIS
\usepackage{aaai25}  % DO NOT CHANGE THIS
\usepackage{times}  % DO NOT CHANGE THIS
\usepackage{helvet}  % DO NOT CHANGE THIS
\usepackage{courier}  % DO NOT CHANGE THIS
\usepackage[hyphens]{url}  % DO NOT CHANGE THIS
\usepackage{graphicx} % DO NOT CHANGE THIS
\usepackage{natbib}  % DO NOT CHANGE THIS AND DO NOT ADD ANY OPTIONS TO IT
\usepackage{caption} % DO NOT CHANGE THIS AND DO NOT ADD ANY OPTIONS TO IT
\usepackage{algorithm}
\usepackage{algorithmic}

\usepackage{newfloat}
\usepackage{listings}
\DeclareCaptionStyle{ruled}{labelfont=normalfont,labelsep=colon,strut=off} % DO NOT CHANGE THIS
\floatstyle{ruled}
\newfloat{listing}{tb}{lst}{}
\floatname{listing}{Listing}
\usepackage{subfigure}

\usepackage{multirow}
\usepackage{booktabs}
\usepackage{amsmath,amsthm,amssymb,amsfonts}
\usepackage{enumitem}
\title{RhythmMamba: Fast,  Lightweight, and Accurate\\ Remote Physiological Measurement}
\author{
    Bochao Zou\textsuperscript{\rm 1},
    Zizheng Guo\textsuperscript{\rm 1},
    Xiaocheng Hu\textsuperscript{\rm 2},
    Huimin Ma\textsuperscript{\rm 1}\thanks{Corresponding author.}
}
\affiliations{
    \textsuperscript{\rm 1}University of Science and Technology Beijing, Beijing, China\\
    \textsuperscript{\rm 2}China Academy of Electronics and Information Technology, Beijing, China\\
    zoubochao@ustb.edu.cn, guozizheng@xs.ustb.edu.cn, 675342900@qq.com, mhmpub@ustb.edu.cn
}

\usepackage{bibentry}
\begin{document}

\maketitle

\begin{abstract}
Remote photoplethysmography (rPPG) is a method for non-contact measurement of physiological signals from facial videos, holding great potential in various applications such as healthcare, affective computing, and anti-spoofing. Existing deep learning methods struggle to address two core issues of rPPG simultaneously: understanding the periodic pattern of rPPG among long contexts and addressing large spatiotemporal redundancy in video segments. These represent a trade-off between computational complexity and the ability to capture long-range dependencies. In this paper, we introduce RhythmMamba, a state space model-based method that captures long-range dependencies while maintaining linear complexity. By viewing rPPG as a time series task through the proposed frame stem, the periodic variations in pulse waves are modeled as state transitions. Additionally, we design multi-temporal constraint and frequency domain feed-forward, both aligned with the characteristics of rPPG time series, to improve the learning capacity of Mamba for rPPG signals.
Extensive experiments show that RhythmMamba achieves state-of-the-art performance with 319\% throughput and 23\% peak GPU memory. 
% The codes are available at https://github.com/zizheng-guo/RhythmMamba.

\end{abstract}

\begin{links}
\link{Code}{https://github.com/zizheng-guo/RhythmMamba}
\end{links}

\section{Introduction}
\label{sec:intro}

Blood Volume Pulse (BVP) is a vital physiological signal, further enabling the extraction of key signs such as heart rate (HR) and heart rate variability (HRV). Photoplethysmography (PPG) is a non-invasive monitoring method that utilizes optical means to measure changes in blood volume within living tissues. 
The physiological mechanism of PPG stems from variations in blood volume during cardiac contraction and relaxation in subcutaneous blood vessels, leading to changes in light absorption and scattering. 
These changes result in periodic color signal variations on imaging sensors, which are imperceptible to the human eye \cite{GREEN,deepphys}. Traditionally, BVP extraction requires the use of contact sensors, which brings inconvenience and limitations. In recent years, non-contact methods for obtaining BVP, particularly rPPG, have garnered increasing attention \cite{review1,AAAI23,aaai24}. 

Early rPPG research primarily relied on traditional signal processing methods to recover weak rPPG signals from facial videos, which are susceptible to interference from environmental light, motion, and other noises. In complex environments, relying solely on signal processing methods often struggles to achieve satisfactory accuracy.
In recent years, data-driven methods have become mainstream, represented by convolutional neural networks (CNNs) and transformers. However, CNNs have limited receptive fields and transformer-based architectures exhibit mediocre performance in capturing long-term dependencies from the computational complexity perspective, especially when dealing with long video sequences.

Recently, Mamba \cite{mamba2} has emerged with its selective state space model, striking a balance between maintaining linear complexity and facilitating long-term dependency modeling. It has been successfully applied to various artificial intelligence tasks such as video understanding \cite{videomamba}. For rPPG tasks that typically require long-term monitoring and are suitable for deployment on mobile devices, the linear complexity and ability to capture long-term dependencies give Mamba an advantage.

However, the direct application of Mamba to rPPG tasks performs poorly. Our experiments reveal that embedding spatiotemporal information into token sequences through patch embedding leads to spatial information significantly disrupting Mamba's comprehension of temporal information (see Section~\ref{sec/4.5}). 
This phenomenon may stem from Mamba's linear modeling characteristics, where the states are associated with the temporal phases of the rPPG signals.
The incorporation of spatial information increases the dimensionality of the state transition process, thereby adding complexity and impeding the model's learning efficacy. 

\begin{figure}[t] 
  \centering 
  \setlength{\abovecaptionskip}{0.07cm}
  \includegraphics[width=0.96\linewidth]{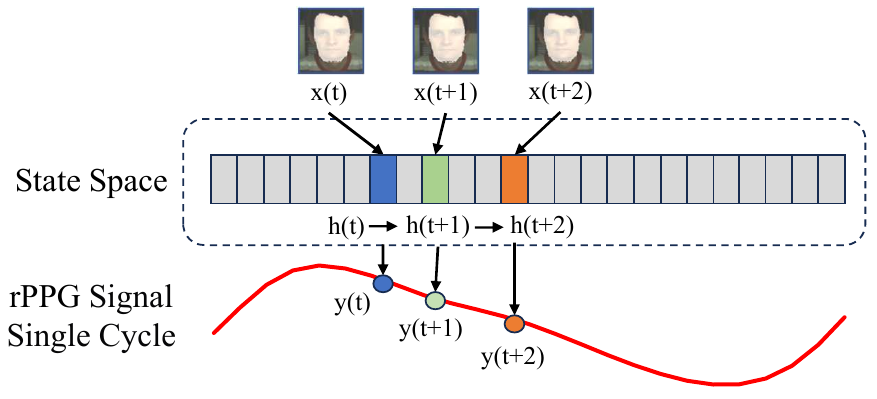}  
  \caption{A schematic diagram of state transitions. Considering the periodic nature of rPPG, the rPPG signal can be represented using a finite number of states. Where $h(t)$ represents the state vector, $x(t)$ represents the input vector, and $y(t)$ represents the output vector.}
  \label{fig:1}
\end{figure}

Motivated by the aforementioned discussion, we propose RhythmMamba, a state space model-based architecture for remote physiological measurement. The proposed frame stem embeds spatial information from a single frame into the channels, thereby viewing the rPPG task as a time series task, allowing the periodic variations in pulse waves to be modeled as state transitions.
As shown in Fig.~\ref{fig:1}, considering the linear modeling characteristics of Mamba, the phase shifts of rPPG signals can be viewed as state transitions within the state space. The periodic nature of rPPG allows the signal to be represented using a finite set of states.

Additionally, we design multi-temporal constraint and frequency domain feed-forward, both aligned with the characteristics of rPPG time series, to improve the learning capacity of Mamba for rPPG signals. By learning from the same sequences with varying temporal lengths, a single Mamba block can simultaneously be constrained by the periodicity of long sequences and the trends of short sequences. Subsequently, through frequency domain feed-forward, the learned temporal features by Mamba undergo inter-channel spatial interaction in the frequency domain, enabling a better discernment of the periodic nature of rPPG signals. 

The main contributions are as follows:

$\bullet$ We propose RhythmMamba, which leverages state space models to model periodic variations as state transitions, combining multi-temporal constraints Mamba and frequency domain feed-forward to learn the quasi-periodic patterns of rPPG. To the best of our knowledge, this is the first work to investigate state space models in the rPPG domain.

$\bullet$ In response to the observed phenomenon where spatial information interferes with Mamba's understanding of temporal sequences, we design the frame stem to embed spatial information into channels, reducing the dimensionality of state transitions to boost Mamba's learning.

\begin{figure}[t] 
  \centering  
  \setlength{\abovecaptionskip}{0.12cm}
  \includegraphics[width=0.9\linewidth]{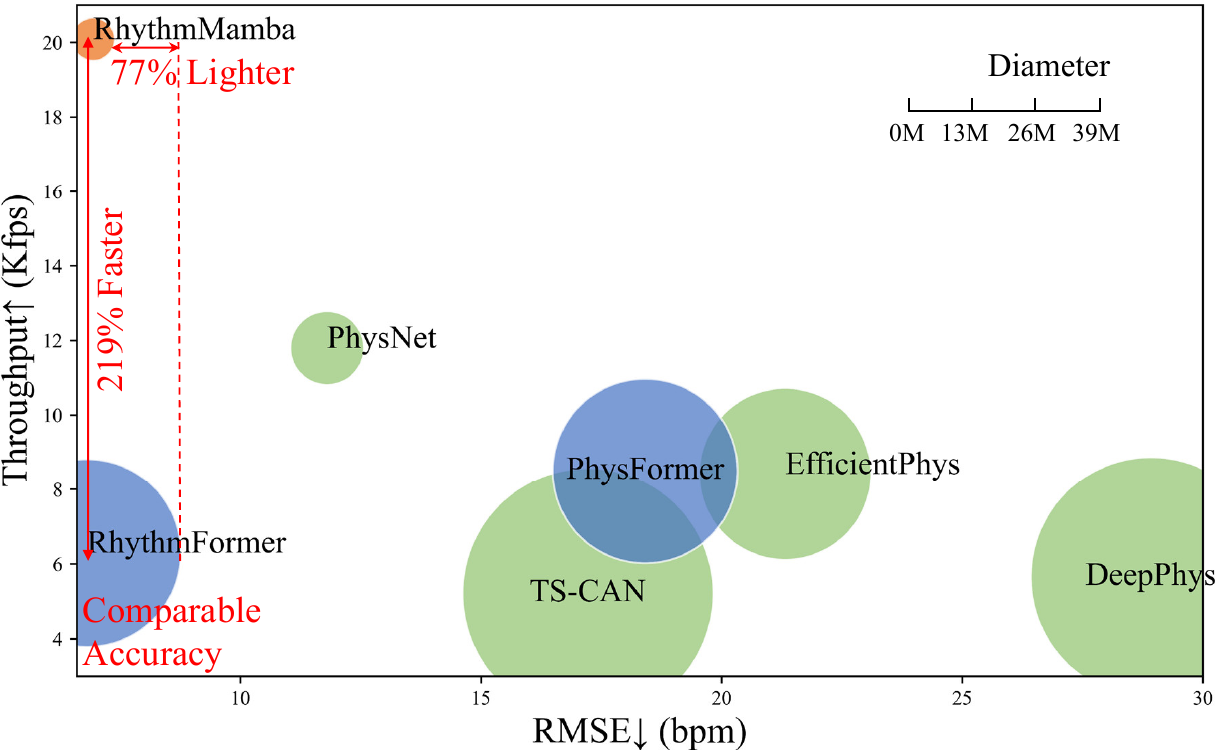}  
  \caption{Performance and efficiency evaluation for intra-dataset testing on MMPD. The diameter of the circle indicates the peak GPU memory. The proposed RhythmMamba is faster, lighter, and achieves comparable accuracy, with these advantages becoming more pronounced as the scale increases due to its linear complexity.}
  \label{fig:2}
\end{figure}

$\bullet$ We conduct extensive experiments on intra-dataset and cross-dataset scenarios. The results demonstrate that RhythmMamba achieves state-of-the-art performance with  319\% throughput and 23\% GPU memory, as illustrated in Fig. \ref{fig:2}.
%------------------------------------------------------------------------

\section{Related Work}
\label{sec:relatedwork}

\subsection{Remote Physiological Measurement}

Early research on rPPG primarily relied on traditional signal processing methods to recover weak rPPG signals from facial videos \cite{GREEN, ICA, CHROM, POS}. In recent years, data-driven approaches have dominated due to their remarkable performance, showcasing a trend in the transition of backbone from 2D CNNs \cite{HR-CNN, synrhythm, deepphys, CVD, TSCAN} to 3D CNNs \cite{physnet, rppgnet, 3DCDC, AAAI23} and further to transformers \cite{physformer, physformer++, efficientphys, tranphys,liu2024spikingphysformer,rhythmformer}. However, none of them have been able to effectively address the two core issues of rPPG: understanding the periodic pattern of rPPG among long contexts and addressing large spatiotemporal redundancy in video segments. This dilemma underscores a trade-off between computational complexity and the ability to capture long-range dependencies, thereby presenting a barrier to deploying rPPG solutions on mobile devices. Although previously dominant 3D CNNs and video transformers have effectively tackled one of the above issues by utilizing local convolutions or long-range attention, they fail to address both problems simultaneously. Unlike them, the proposed RhythmMamba can capture long-range dependencies while maintaining linear complexity, making it fast, lightweight, and accurate for remote physiological measurement.

\subsection{Vision Mamba}

Recently, Mamba has distinguished itself with a data-dependent state space model (SSM) and a selection mechanism utilizing parallel scanning, striking a balance between maintaining linear complexity and facilitating long-term dependency modeling. Compared to transformers with quadratic complexity attention \cite{transformer,vivit}, Mamba excels at handling long sequences with linear complexity. Subsequently, the immense potential of Mamba has sparked a series of works \cite{visionmamba,simba,videomamba}, demonstrating superior performance and higher GPU efficiency of Mamba over Transformers on downstream vision tasks. 
However, unlike other video tasks, rPPG signals are particularly weak and highly susceptible to noise from factors such as lighting and motion, making the direct application of the traditional Mamba architecture to rPPG tasks perform poorly. 
In contrast to previous works, we view the rPPG task as the time series task, fully integrating spatial information into the channels and designing multiple modules tailored for time series to boost Mamba's learning.

\label{sec/3_methodology}

\begin{figure*}[t] 
  \centering  
  \includegraphics[width=0.87\linewidth]{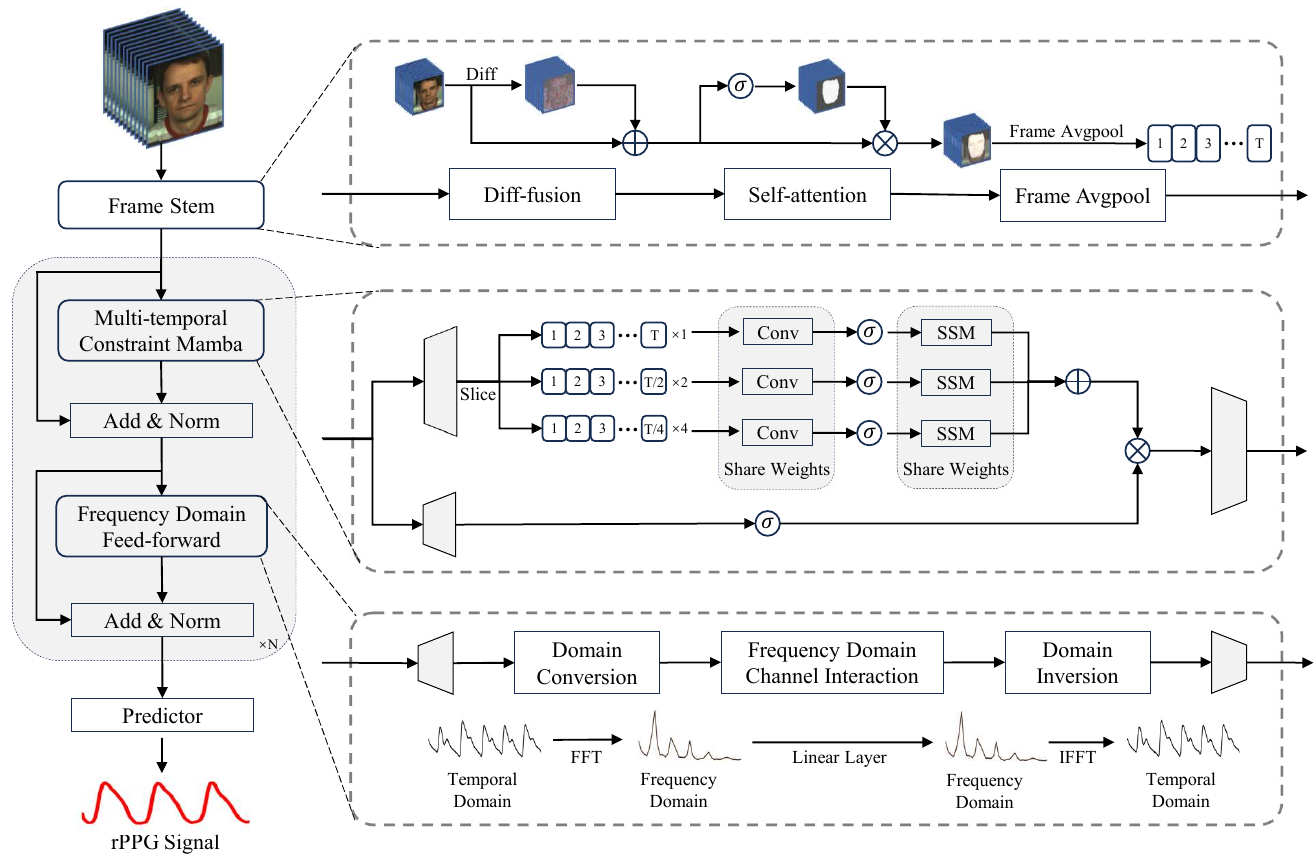}  
  \caption{The framework of RhythmMamba. It consists of frame stem, multi-temporal constraint Mamba, frequency domain feed-forward, and rPPG predictor head. Where "$+$" represents addition, "$\times$" represents multiplication, "$\sigma$" represents the activation layer, and trapezoid represents the linear layer.}
  \label{fig:main}
\end{figure*}

\section{Methodology}

Section \ref{sec/3.1} introduces the general framework of RhythmMamba, followed by the presentation of its main components: the frame stem in Section \ref{sec/3.2}, the multi-temporal constraint Mamba in Section \ref{sec/3.3}, and lastly, the frequency domain feed-forward in Section \ref{sec/3.4}.

\subsection{The General Framework of RhythmMamba}
\label{sec/3.1}

As shown in Figure \ref{fig:main}, RhythmMamba consists of frame stem, multi-temporal constraint Mamba, frequency domain feed-forward, and rPPG predictor head. The frame stem utilizes diff-fusion, self-attention, and frame average pooling to extract rPPG features and embed all spatial information into channels. Specifically, given an RGB video input $X \in \mathbb{R}^{3 \times T \times H \times W}$, $X_{stem}=frame\_stem(X)$, where $X_{stem} \in \mathbb{R} ^{T\times C}$, and C, T, W, H indicate channel, sequence length, width, and height, respectively.

Subsequently, the output of the frame stem will be fed into the multi-temporal constraint Mamba. The tokens will be sliced into sequences of varying temporal lengths, followed by the processing of hidden information between tokens with the SSM. Then, the output of Mamba will be passed into the frequency domain feed-forward, facilitating the interaction of information across multiple channels. The outputs of Mamba and feed-forward network (FFN) will undergo normalization and residual connections. The two outputs have dimensions identical to the output of the frame stem $X_{stem} \in \mathbb{R} ^{T\times C}$. Finally, the rPPG features will be projected into PPG waves through the predictor head.

\subsection{Frame Stem}
\label{sec/3.2}

In the field of video understanding, existing transformer-based and Mamba-based methods typically embed spatiotemporal information into token sequences through patch embedding \cite{vivit,visionmamba,vit}. Previous works on rPPG have also been based on such foundational models for improvements. For transformer-based methods, spatiotemporal token sequences can inspire long-range spatiotemporal attention both within frames and across frames. However, we found that for linear Mamba, spatial information may interfere with Mamba's understanding of temporal sequences [see section \ref{sec/4.5}]. The frame stem is utilized to initially extract rPPG features and embed spatial information fully into the channels, thereby boosting the learning of state transitions in the multi-temporal constraint Mamba and the channel interactions in the frequency domain feed-forward.

Firstly, the diff-fusion module \cite{rhythmformer} integrates frame differences into the raw frames, enabling frame-level representation awareness of BVP wave variations. This effectively enhances the features of rPPG with a small additional computational cost. Additionally, for rPPG, high-frequency information across frames and low-frequency information within frames are required. Therefore, relatively large convolutional kernels are used to obtain low-frequency information within frames, ensuring that spatial information is fully incorporated into the channels. Here, 'relatively large' refers to the size relative to the image resolution, enabling a large receptive field.

Specifically, for an RGB video input $X \in \mathbb{R}^{3 \times T \times H \times W}$, temporal shift is initially applied to obtain $X_{t-2}$, $X_{t-1}$, $X_t$, $X_{t+1}$ and $X_{t+2}$. Subsequently, frame differences between consecutive frames are computed in reverse chronological order, yielding $D_{t-2}$, $D_{t-1}$, $D_{t+1}$, and $D_{t+2}$. The frame differences and the raw frames are then passed through ${Stem}_1$ for feature extraction. ${Stem}_1$ consists of a 2D convolution layer with $(7\times7)$ kernel, followed by batch normalization (BN), ReLU, and MaxPool. The input dimension is 3 when taking raw frames as input, and 12 when taking the concatenation of frame differences as input.
\begin{small}
\begin{equation}
\begin{split}
  &X_{diff}\ =Stem_1(Concat\left(D_{t-2},D_{t-1},D_{t+1},D_{t+2}\right)),\\
  &X_{raw}\ =Stem_1(X_t).
\end{split}
\end{equation}
\end{small}

Then frame differences $X_{diff}$ and raw frames $X_{raw}$ are merged and the feature representation is further enhanced through ${Stem}_2$, which consists of a 2D convolution layer with $(7\times7)$ kernel, followed by BN, ReLU and MaxPool.
\begin{small}
\begin{equation}
% \small
  X_{fusion}= {Stem}_2(X_{raw} + X_{diff}) + {Stem}_2(X_{diff}).
\end{equation}
\end{small}

Subsequently, at the resolution of $(16\times16)$, ${Stem}_3$ utilizes a convolution layer with $(5\times5)$ kernel to fully integrate spatial information into the channels, followed by BN. Before frame-level global average pooling, a self-attention module is employed to enhance skin regions with rPPG signals in the spatial domain. This self-attention module utilizes sigmoid activation followed by L1 normalization, which is softer than softmax and generates fewer masks \cite{efficientphys}. The attention mask can be computed as:
\begin{small}
\begin{equation}
% \small
  Mask=\frac{(H/8)(W/8) \cdot \sigma(Stem_3(X_{fusion}))}{2||\sigma(Stem_3(X_{fusion}))||_1}.
\end{equation}
\end{small}

Finally, the attention output $X_{attn}\in \mathbb{R}^{C \times T \times H/8 \times W/8}$, undergoes global average pooling within each frame, resulting in the stem output $X_{stem}\in \mathbb{R}^{T\times C}$.

\subsection{Multi-temporal Constraint Mamba}
\label{sec/3.3}
Previous studies \cite{mul1,mul2,mul3} have shown the effectiveness of modeling periodic tasks with multi-temporal scales, primarily achieved by the extraction and fusion of features from different temporal scales.
Unlike these studies, we replace multi-temporal fusion with multi-temporal constraint, which better aligns with the characteristics of the Mamba. Specifically, we slice a video segment into numerous sub-segments of varying lengths to constrain a single Mamba block, rather than downsampling the video segment to different resolutions and using multiple Mamba blocks to extract and fuse features.
The aim is to subject a Mamba block to both the periodic constraints of long sequences and the trend constraints of short sequences simultaneously, instead of extracting different features from multi-temporal scales.

After the frame stem, the token sequence can be regarded as a time series, and the state transition of Mamba can be interpreted as the temporal phase shift. Due to the quasi-periodicity of rPPG signals, the signal can be represented using a finite set of states.
Specifically, as illustrated by the multi-temporal constraint Mamba in Figure \ref{fig:main}, the input $X_{stem}$ is first linearly projected and then processed through three weight-shared paths. Along these paths, the input is sliced into temporal sequences of varying lengths.
For the $i_{th}$ path, the sequence is divided into $2^{i-1}$ sub-sequences, each of which undergoes sequential processing through a convolution layer, activation layer, and selective state space model (see supplementary material A for details). Subsequently, they are recombined into a sequence of the original length, forming the output of the $i_{th}$ path, denoted as $X_{path_i}$. The output before projection can be represented as follows:
% \begin{small}
\begin{equation}
  X_{mamba}= \sum\limits_{i=1}^{3}X_{path_i} \times \sigma(Proj(X_{stem})).
\end{equation}
% \end{small}

\subsection{Frequency Domain Feed-forward}
\label{sec/3.4}
The FFN employs linear transformations to project data into a higher-dimensional space before mapping it back into a lower-dimensional space. Through this channel interaction, deeper features are extracted. 
In previous rPPG studies, spatio-temporal FFN was frequently applied, which introduced a depthwise 3D convolution layer between the two linear layers of the vanilla FFN, to refine the local inconsistency and provide relative positional cues. \cite{physformer}. 

In our study, due to the input sequences being solely time-dependent, channel interaction in the frequency domain enables a better discernment of the periodic nature of rPPG signals. So we introduce frequency domain feed-forward, which adds a frequency domain linear layer between the two linear layers of the vanilla FFN. The frequency domain linear layer consists of three stages: domain conversion, frequency domain channel interaction, and domain inversion.

\textbf{Domain Conversion/Inversion.} 
Domain conversion and inversion utilize fast Fourier transform and inverse Fourier transform, respectively. The utilization of the Fourier transform enables the decomposition of rPPG signals into their constituent frequencies, facilitating the recognition of periodic patterns in the rPPG signals. We transform the input $H(t)$ to the frequency domain $H(f)$ as follows:
\begin{small}
\begin{equation}
\begin{split}
H(f)& = \int_{-\infty}^{\infty} H(t) e^ {-j2\pi ft}dt \\
 = &\int_{-\infty}^{\infty} H(t) \cos(2\pi ft)dt + j\int_{-\infty}^{\infty} H(t) \sin(2\pi ft)dt \\
 = &{H(f)}_{re} + j{H(f)}_{im}.
\end{split}
\end{equation}
\end{small}

\begin{table*}[t]
  \centering
  \small
  \renewcommand\arraystretch{0.96}
  \setlength{\abovecaptionskip}{0.1cm}
    \begin{tabular}{ccccccccccccc}
    \toprule
          & \multicolumn{3}{c}{PURE} & \multicolumn{3}{c}{UBFC} & \multicolumn{3}{c}{VIPL-HR} & \multicolumn{3}{c}{MMPD} \\
\cmidrule{2-13}    Method & MAE$\downarrow$ & RMSE$\downarrow$ & $\rho\uparrow$ &  MAE$\downarrow$ & RMSE$\downarrow$ & $\rho\uparrow$ & MAE$\downarrow$ & RMSE$\downarrow$ & $\rho\uparrow$ &  MAE$\downarrow$ & RMSE$\downarrow$ & $\rho\uparrow$\\
    \midrule
    HR-CNN  & 1.84  & 2.37  & 0.98  & 4.90  & 5.89  & 0.64  & -     & -     & -     & -     & -     & - \\
    DeepPhys  & 0.83  & 1.54  & 0.99  & 6.27  & 10.82 & 0.65  & 11.00 & 13.80 & 0.11  & 22.27 & 28.92 & -0.03 \\
    PhysNet  & 2.10  & 2.60  & 0.99  & 2.95  & 3.67  & 0.97  & 10.80 & 14.80 & 0.20  & 4.80  & 11.80 & 0.60 \\
    TS-CAN  & 2.48  & 9.01  & 0.92  & 1.70  & 2.72  & 0.99  & -     & -     & -     & 9.71  & 17.22 & 0.44 \\
    Gideon et al. & 2.30  & 2.90  & 0.99  & 1.85  & 4.28  & 0.93  & 9.01  & 14.02 & 0.58  & -     & -     & - \\
    Dual-GAN  & 0.82  & 1.31  & 0.99  & \underline{0.44}  & \underline{0.67}  & 0.99  & 4.93  & 7.68  & \underline{0.81}  & -     & -     & - \\
    PhysFormer  & 1.10  & 1.75  & 0.99  & 0.50  & 0.71  & 0.99  & 4.97  & 7.79  & 0.78  & 11.99 & 18.41 & 0.18 \\
    EfficientPhys  &  -      &  -      &  -      & 1.14  & 1.81  & 0.99  & -     & -     & -     & 13.47 & 21.32 & 0.21 \\
    TFA-PFE   & 1.44  & 2.50  &  -      & 0.76  & 1.62  &  -    & -     & -     & -     & -     & -     & - \\
    NEST  & -     & -     & -     & -     & -     & -     & \underline{4.76}  & \underline{7.51}  & \textbf{0.84} & -     & -     & - \\
    Li et al.   & 0.64  & 1.16  & 0.99  & 0.48  & \textbf{0.64} & 0.99  & 5.19  & 8.26  & 0.78  & -     & -     & - \\
    Yue et al.   & 1.23  & 2.01  & 0.99  & 0.58  & 0.94  & 0.99  & -     & -     & -     & -     & -     & - \\
    PhysFormer++ & -     & -     & -     & -     & -     & -     & 4.88  & 7.62  & 0.80  & -     & -     & - \\
    Contrast-Phys+   & 0.48  & 0.98  & 0.99  & \textbf{0.21} & 0.80  & 0.99  & -     & -     & -     & -     & -     & - \\
    RhythmFormer   & \underline{0.27}  & \underline{0.47}  & 0.99  & 0.50  & 0.78  & 0.99  & -  & - & - & \textbf{3.07}  & \textbf{6.81} & \textbf{0.86} \\
    Ours   & \textbf{0.23} & \textbf{0.34} & 0.99  & 0.50  & 0.75  & 0.99  & \textbf{4.30} & \textbf{7.49} & \underline{0.81}  & \underline{3.16} & \underline{7.27}  & \underline{0.84} \\
    \bottomrule
    \end{tabular}%
    \caption{Intra-dataset evaluation. Best results are marked in \textbf{bold} and second best in \underline{underline}.}
  \label{tab:intra}%
\end{table*}%

Where $f$ represents frequency, $t$ represents time, and the subscripts $re$ and $im$ denote the real and imaginary components of the corresponding complex data, respectively. After channel Interaction in the frequency domain, inverse Fourier transform is employed to revert to the temporal domain:
\begin{small}
\begin{equation}
\begin{split}
% \small
H(t)&=  \int_{-\infty}^{\infty} H(f) e^ {j2\pi ft}df \\
&= \int_{-\infty}^{\infty}(H(f)_{re} + jH(f)_{im})e^{j2\pi ft}df.
\end{split}
\end{equation}
\end{small}

\textbf{Frequency Domain Channel Interaction.} 
Through the frame stem, spatial information is embedded into the channels, each of which is treated as a time series. Consequently, the frequency domain features obtained after domain conversion can clearly represent the signal's frequency composition. This allows channel interactions in the frequency domain to refine noise interference and more easily focus on critical channels.
Specifically, channel interaction is implemented through a linear layer, the theoretical feasibility of which has been demonstrated by \cite{fremlp}. For complex input $H\in \mathbb{R}^{T\times C}$, given complex weight matrix $W\in \mathbb{R}^{C\times C}$ and complex bias $B\in \mathbb{R}^{C}$, according to the rules of complex multiplication, it can be expressed as:
\begin{small}
\begin{equation}
\begin{split}
&H'_{re} = H_{re}W_{re} - H_{im}W_{im} + B_{re},\\
&H'_{im} = H_{re}W_{im} + H_{im}W_{re} + B_{im},\\
&H' = H'_{re} + j\cdot H'_{im}.
\end{split}
\end{equation}
\end{small}

After inverse FFT transform, the frequency domain feed-forward outputs through linear projection, resulting in $X_{FFN} \in \mathbb{R} ^{T\times C}$.

\label{sec/4_experiment}
\section{Experiment}

\subsection{Dataset and Performance Metric}
The experiments of remote physiological measurement were conducted on four publicly available datasets: PURE~\cite{PURE}, UBFC-rPPG~\cite{UBFC-rPPG}, VIPL-HR~\cite{rhythmNet}, and MMPD~\cite{MMPD}. \textbf{PURE} comprises 59 1-minute videos, documenting records of 10 subjects, each engaging in six different activities. \textbf{UBFC-rPPG} consists of 42 videos, recording 42 subjects. These videos were derived from a setting where subjects participated in a time-limited digital game. \textbf{VIPL-HR} includes 2,378 RGB videos from 107 participants, captured using three RGB cameras, with an unstable fps. \textbf{MMPD} includes 660 1-minute videos, documenting records of 33 subjects. Participants engaged in four different activities under four distinct lighting conditions. 
\textbf{Metircs.} The evaluation was conducted using five metrics for video-level heart rate estimations: Mean Absolute Error (MAE), Root Mean Square Error (RMSE), Mean Absolute Percentage Error (MAPE), Pearson Correlation Coefficient ($\rho$), and Signal-to-Noise Ratio (SNR).

\begin{table*}[ht]
  \small
  \centering
  \renewcommand\arraystretch{0.96}
  \setlength{\abovecaptionskip}{0.12cm}
  \setlength{\tabcolsep}{2.8pt}
    \begin{tabular}{ccccccc|ccccc|ccccc}
    \toprule
          &       & \multicolumn{15}{c}{Test Set} \\
\cmidrule{3-17}          &       & \multicolumn{5}{c}{PURE}              & \multicolumn{5}{c}{UBFC}         & \multicolumn{5}{c}{MMPD} \\
\cmidrule{3-17}    Method & TrainSet & MAE   & RMSE  & MAPE  & $\rho$ & \multicolumn{1}{c}{SNR} & MAE   & RMSE  & MAPE  & $\rho$ & \multicolumn{1}{c}{SNR} & MAE & RMSE & MAPE & $\rho$ & SNR \\
    \midrule
    GREEN & -   & 10.09 & 23.85 & 10.28 & 0.34  & -2.66 & 19.73 & 31.00 & 18.72 & 0.37  & -11.18 & 21.68 & 27.69 & 24.39 & -0.01 & -14.34 \\
    ICA   & -   & 4.77  & 16.07 & 4.47  & 0.72  & 5.24  & 16.00 & 25.65 & 15.35 & 0.44  & -9.91 & 18.60 & 24.30 & 20.88 & 0.01  & -13.84 \\
    CHROM & -   & 5.77  & 14.93 & 11.52 & 0.81  & 4.58  & 4.06  & 8.83  & 3.84  & 0.89  & -2.96 & 13.66 & 18.76 & 16.00 & 0.08  & -11.74 \\
    LGI   & -   & 4.61  & 15.38 & 4.96  & 0.77  & 4.50  & 15.80 & 28.55 & 14.70 & 0.36  & -8.15 & 17.08 & 23.32 & 18.98 & 0.04  & -13.15 \\
    PBV   & -   & 3.92  & 12.99 & 4.84  & 0.84  & 2.30  & 15.90 & 26.40 & 15.17 & 0.48  & -9.16 & 17.95 & 23.58 & 20.18 & 0.09  & -13.88 \\
    POS   & -   & 3.67  & 11.82 & 7.25  & 0.88  & 6.87  & 4.08  & 7.72  & 3.93  & 0.92  & -2.39 & 12.36 & 17.71 & 14.43 & 0.18  & -11.53 \\
    OMIT & - & 4.66 & 15.82 & 4.97 & 8.76 & 4.37 & 16.99  & 29.54 & 15.91 & 0.34 & -7.29  & 17.02 & 23.23 & 18.89 & 0.04 & -12.77 \\
    \midrule
    \multirow{2}[0]{*}{DeepPhys} & UBFC & 5.54  & 18.51 & 5.32  & 0.66  & 4.40  & -   & -   & -   & -   & -   & 17.50 & 25.00 & 19.27 & 0.06  & -11.72 \\
          & PURE  & -   & -   & -   & -   & -   & 1.21  & 2.90  & 1.42  & 0.99  & 1.74  & 16.92 & 24.61 & 18.54 & 0.05  & -11.53 \\
    \multirow{2}[0]{*}{PhysNet} & UBFC & 8.06  & 19.71 & 13.67 & 0.61  & 6.68  & -   & -   & -   & -   & -   & \underline{9.47} & \underline{16.01} & \textbf{11.11} & 0.31  & \underline{-8.15} \\
          & PURE  & -   & -   & -   & -   & -   & 0.98  & 2.48  & 1.12  & 0.99  & 1.49 & 13.94 & 21.61 & 15.15 & 0.20  & -9.94 \\
    \multirow{2}[1]{*}{TS-CAN} & UBFC & 3.69  & 13.80 & 3.39  & 0.82  & 5.26  & -   & -   & -   & -   & -   & 14.01 & 21.04 & 15.48 & 0.24  & -10.18 \\
          & PURE  & -   & -   & -   & -   & -   & 1.30  & 2.87  & 1.50  & 0.99  & 1.49  & 13.94 & 21.61 & 15.15 & 0.20  & -9.94 \\
    \multirow{2}[0]{*}{PhysFormer} & UBFC & 12.92  & 24.36  & 23.92  & 0.47  & 2.16 & -   & -   & -   & -   & -   & 12.10 & 17.79 & 15.41 & 0.17  & -10.53 \\
          & PURE  & -   & -   & -   & -   & -   & 1.44  & 3.77  & 1.66  & 0.98  & 0.18  & 14.57 & 20.71 & 16.73 & 0.15  & -12.15 \\
    \multirow{2}[0]{*}{EfficientPhys} & UBFC & 5.47  & 17.04 & 5.40  & 0.71  & 4.09  & -   & -   & -   & -   & -   & 13.78 & 22.25 & 15.15 & 0.09  & -9.13 \\
          & PURE  & -   & -   & -   & -   & -   & 2.07  & 6.32  & 2.10  & 0.94  & -0.12 & 14.03 & 21.62 & 15.32 & 0.17  & -9.95 \\
    \multirow{2}[0]{*}{Spiking-Phys.} & UBFC  & 3.83  & -     & 5.70  & 0.83  & -     & -     & -     & -     & -     & -     & 14.15 & -     & 16.22 & 0.15  & - \\
          & PURE  & -     & -     & -     & -     & -     & 2.80  & -     & 2.81  & 0.95  & -     & 14.57 & -     & 16.55 & 0.14  & - \\
    \multirow{2}[1]{*}{RhythmFormer} & UBFC & \textbf{0.97} & \textbf{3.36} & \textbf{1.60} & \textbf{0.99} & \textbf{12.01} & -   & -   & - & - & - & \textbf{9.08} & \textbf{15.07} & \underline{11.17} & \textbf{0.53} & \textbf{-7.73} \\
          & PURE  & -   & -   & -   & -   & -   & \textbf{0.89} & \textbf{1.83} & \textbf{0.97} & \textbf{0.99} & \underline{6.05} & \textbf{8.98} & \textbf{14.85} & \textbf{11.11} & \textbf{0.51} & \underline{-8.39} \\
    \multirow{2}[1]{*}{Ours} & UBFC & \underline{1.98} & \underline{6.51} & \underline{3.59} & \underline{0.96} & \underline{8.94} & -   & -   & - & - & - & 10.63 & 17.14 & 12.14 & \underline{0.34} & -8.28 \\
          & PURE  & -   & -   & -   & -   & -   & \underline{0.95} & \textbf{1.83} & \underline{1.04} & \textbf{0.99} & \textbf{6.35} & \underline{10.44} & \underline{16.70} & \underline{12.25} & \underline{0.36} & \textbf{-8.18} \\
    \bottomrule
    \end{tabular}%
  \caption{Cross-dataset evaluation. Best results are marked in \textbf{bold} and second best in \underline{underline}.}
  \label{tab:cross}%
\end{table*}

\subsection{Implementation Details}
The proposed RhythmMamba was implemented based on PyTorch, and we utilized an open-source rPPG toolbox \cite{toolbox} to conduct a fair comparison against several state-of-the-art methods.
In the pre-processing, video inputs were divided into segments of 160 frames. Facial recognition was applied on the first frame of each segment, followed by cropping and resizing of the facial region. These adjustments were then maintained throughout the subsequent frames.
In the post-processing, a second-order Butterworth filter (cutoff frequencies: 0.75 and 2.5 Hz) was applied to filter the rPPG waveform, and power spectral density was computed by the Welch algorithm for further heart rate estimation. Following the protocol outlined in \cite{autohr}, random upsampling, downsampling, and horizontal flipping were applied for data augmentation. The experiment was conducted on NVIDIA RTX 3090.

\textbf{Loss.} We employed a loss function that integrates constraints from both the temporal and frequency domains \cite{autohr}. 
The negative Pearson correlation coefficient is utilized as temporal constraint $\mathcal{L}_{Time}$, while cross-entropy between the power spectral density of prediction and the HR derived from the power spectral density of ground truth, is employed as frequency constraint $\mathcal{L}_{Freq}$.
\begin{small}$\mathcal{L}_{Freq}=CE(maxIndex(PSD(PPG_{gt})),PSD(PPG_{pred}))$\end{small}, where $PSD$ represents Power Spectral Density and $maxIndex$ represents the index of the maximum value. The overall loss is expressed by: $\mathcal{L}_{overall}=a\cdot\mathcal{L}_{Time}+b\cdot\mathcal{L}_{Freq}$.

\textbf{Comparison.} We compare our method with state-of-the-art approaches in intra-dataset testing \cite{HR-CNN,deepphys,physnet,TSCAN,gideon2021iccv,dualgan,physformer,efficientphys,AAAI23,nest,liiccv23,yue2023,physformer++,contrastphys+,rhythmformer}. Building on this, additional comparisons with \cite{GREEN,ICA,CHROM,LGI,PBV,POS,casado2023face2ppg,liu2024spikingphysformer} are presented in cross-dataset testing.

\subsection{Intra-Dataset Evaluation}
We conducted intra-dataset evaluation on the PURE and UBFC datasets to validate the feasibility of the Mamba architecture. For the evaluation of the PURE dataset, we followed the protocols outlined in \cite{dualgan}, splitting the dataset sequentially into training and testing sets with a ratio of 6:4. Similarly, for the evaluation of the UBFC dataset, we followed the protocols in \cite{dualgan}, selecting the first 30 samples as the training set and the remaining 12 samples as the testing set. Due to the absence of the validation set, we selected the checkpoint from the last epoch for testing and compared them with the reported results from previous methods. As shown in Table \ref{tab:intra}, on the PURE dataset, our method outperformed all state-of-the-art methods across all metrics, achieving the minimum MAE (0.23) and RMSE (0.34). On the UBFC dataset, our method also achieved comparable performance to others.

Due to the relative simplicity of PURE and UBFC, the performance of state-of-the-art methods on these datasets is nearing saturation. To further evaluate the performance, we employed the more challenging dataset. For the VIPL-HR dataset, we followed the subject-exclusive 5-fold cross-validation protocol, as outlined in \cite{rhythmNet, physformer}. For the MMPD dataset, following the protocols outlined in \cite{rhythmformer}, the dataset was sequentially split into training, validation, and testing sets with a ratio of 7:1:2. As shown in Table \ref{tab:intra}, our method achieved comparable performance to the previous methods. This indicates that RhythmMamba can accurately extract weak rPPG signals and understand their periodic nature, which provides ample empirical evidence for the feasibility of the Mamba architecture in the rPPG task.

\begin{table*}[ht]
  \small
  \centering
  \setlength{\abovecaptionskip}{0.1cm}
  \setlength{\tabcolsep}{6.5pt}
    \begin{tabular}{cccccccccc}
    \toprule
    Diff-fusion & Self-attention &  Large Kernel & Multi-temporal & FFN   & MAE$\downarrow$ & RMSE$\downarrow$ & MAPE$\downarrow$ & $\rho\uparrow$ & SNR$\uparrow$  \\
    \midrule
    × & \checkmark  & \checkmark & \checkmark & Frequency & 5.71 & 10.00 & 5.97 & 0.68 & -0.29 \\
    \checkmark &×   & \checkmark & \checkmark & Frequency & 4.02  & 8.43  & 4.15  & 0.79  & 2.98 \\
    \checkmark & \checkmark & ×   & \checkmark & Frequency & 3.51  & 7.53  & 3.81  & 0.83  & 4.22 \\
    \checkmark & \checkmark & \checkmark   & ×   & Frequency & 3.60  & 7.78  & 3.83  & 0.81  & 4.38 \\
    \checkmark & \checkmark  & \checkmark   & \checkmark   & ×   & 3.53  & 7.65  & 3.67  & 0.83  & 2.89 \\
    \checkmark & \checkmark  & \checkmark   & \checkmark   & Vanilla & 3.54  & 7.68  & 3.72  & 0.82  & 2.88 \\
    \checkmark & \checkmark  & \checkmark   & \checkmark   & Spatio-Temporal & 3.82  & 8.06  & 3.95  & 0.81  & 4.06 \\
    \checkmark & \checkmark  & \checkmark   & \checkmark   & Frequency & \textbf{3.16} & \textbf{7.27} & \textbf{3.37} & \textbf{0.84} & \textbf{4.74} \\
    \bottomrule
    \end{tabular}%
  \caption{Impact of key modules.}
  \label{tab:ablation}%
\end{table*}%

\begin{table}[t]
  \small
  \centering
  \setlength{\abovecaptionskip}{0.1cm}
    \setlength{\tabcolsep}{3.0pt}
    \begin{tabular}{cccccc}
    \toprule
    Spatial Token numbers & MAE$\downarrow$ & RMSE$\downarrow$ & MAPE$\downarrow$ & $\rho\uparrow$ & SNR$\uparrow$  \\
    \midrule
    8×8   & 4.90  & 10.14 & 5.15  & 0.71  & -2.10 \\
    4×4   & 4.62  & 8.87  & 4.83  & 0.77  & -0.88 \\
    4×4 (Temporal Embed) & 4.92  & 10.04 & 5.07  & 0.69  & -1.43 \\
    4×4 (Position Embed) & 5.47  & 10.34 & 5.77  & 0.71  & -3.03 \\
    2×2   & 4.69  & 9.97  & 4.80  & 0.70  & -0.98 \\
    1×1 (Avgpool) & \textbf{3.54}  & \textbf{7.68}  & \textbf{3.72}  & \textbf{0.82}  & \textbf{2.88} \\
    \bottomrule
    \end{tabular}%
  \caption{Impact of spatial information (with vanilla FFN).}
  \label{tab:spatio}%
\end{table}%

\subsection{Cross-Dataset Evaluation}
To objectively evaluate the generalization capability to out-of-distribution data, we followed the protocols outlined in \cite{toolbox} for cross-dataset evaluation. The models were trained on either the PURE or UBFC datasets and tested on the PURE, UBFC, and MMPD datasets. The training dataset was sequentially split into training and validation sets with a ratio of 8:2. All comparative methods were implemented based on the rPPG toolbox \cite{toolbox}. As shown in Table \ref{tab:cross}, the proposed RhythmMamba also achieved SOTA performance, demonstrating its capability in modeling domain-invariant features and generalizing to unseen domains. Based on the comparisons in Tables \ref{tab:intra} and \ref{tab:cross}, the improvement in cross-dataset results is less significant compared to intra-dataset results, possibly due to the fine-grained token-wise self-attention, which may have an advantage in capturing domain-invariant features.
Nevertheless, founded on fewer parameters and lower computational complexity, RhythmMamba showcases its potential in real-world applications through its robustness and generalization in complex environments. Additional visualization results can be found in Supplementary Material B.

\subsection{Ablation Study}
\label{sec/4.5}

Ablation studies were conducted on the MMPD dataset to assess the impact of different modules.

\textbf{Impact of Spatial Information.} As shown in Table \ref{tab:spatio}, the ablation study of spatial information is presented, where both position embedding and temporal embedding were implemented using learnable parameters. For a fair comparison, the vanilla FFN was used, as the frequency domain FFN might have an advantage with purely temporal token sequences.
It is evident that tokenized spatiotemporal information performs poorly, even with temporal embedding or position embedding. The integration of spatial information increases the dimensionality of the state transition process, thereby elevating complexity and making the model more difficult to train.
We view the rPPG task as a time series task, embedding spatial information into the channels, with each channel being treated as a purely temporal sequence.
This ensures that the state transition process occurs purely along the temporal dimension, while spatial information interactions are facilitated through subsequent channel interactions, effectively resolving this issue.

\textbf{Impact of Key Modules.} As illustrated in Table \ref{tab:ablation}, the comparison between the first four rows and the last row indicates the significant roles played by these modules. 
The diff-fusion module enables frame-level representation awareness of BVP wave variations, effectively enhancing rPPG features with a small additional computational cost.
The use of relatively large convolution kernels and self-attention allows for the integration of spatial information into channels effectively, thereby providing sufficient information for subsequent processing.
The multi-temporal constraint Mamba constrains a single Mamba block simultaneously to short-term trends and periodic patterns, facilitating the accurate comprehension of rPPG features.

\textbf{Impact of Frequency Domain Feed-forward.} As shown in Table \ref{tab:ablation}, the last four rows show that the frequency domain FFN plays an important role. Among them, vanilla FFN refers to using two linear layers to compose the FFN, and Spatio-Temporal FFN refers to the addition of a depthwise convolution layer between the linear layers \cite{physformer}. Frequency domain FFN adds a frequency domain linear layer between the linear layers, effectively extracting the most critical frequency domain features in rPPG signals.

\subsection{Computational Cost}
We conducted a 30-second inference test at a resolution of 128×128, reporting the parameters, average MACs per frame, average throughput per frame, and average peak GPU memory usage per frame. As shown in Table \ref{tab:cost} and Figure \ref{fig:2}, RhythmMamba achieved 319\% throughput and 23\% peak GPU memory, demonstrating the potential for effective mobile-level rPPG applications. Building on this advantage, we also found that the proposed method can accept inputs of arbitrary length during inference without any performance degradation (see Supplementary Material C for details).

\begin{table}[t]
  \small
  \centering
  \setlength{\abovecaptionskip}{0.1cm}
    \begin{tabular}{ccccc}
    \toprule
    Method & Para. & MACs & Throughput & Memory \\
    \midrule
    DeepPhys & 1.98  & 744.45 & 5.65  & 37.28 \\
    PhysNet & 0.77  & 438.24 & 11.80 & 11.43 \\
    TS-CAN & 1.98  & 744.45 & 5.21  & 38.91 \\
    PhysFormer & 7.38  & 316.29 & 8.50  & 28.63 \\
    EfficientPhys & 1.91  & 373.72 & 8.42  & 26.68 \\
    RhythmFormer & 3.25  & 240.55 & 6.30  & 29.06 \\
    Ours & 1.07  & 80.90 & 20.09 & 6.66 \\
    \bottomrule
    \end{tabular}%
  \caption{Computational cost. The horizontal axis represents Parameters (M), MACs (M), Throughput (Kfps), and Peak GPU Memory Usage (M).}
  \label{tab:cost}%
\end{table}%

\label{sec/5_conclusion}
\section{Conclusion}

We approach the rPPG task as a time series task, designing multiple modules that align with the temporal characteristics of rPPG signals to boost state space model learning. This approach boasts strong long-range dependency modeling capabilities while maintaining linear complexity.
It achieves state-of-the-art performance both within and across datasets with a faster and more lightweight design. However, since Mamba's state transitions align closely with the periodic variations of rPPG signals, we believe that Mamba's potential in rPPG extends beyond the current results. Our utilization of periodic priors is currently limited and we would like to delve into this more deeply in the future.

\section*{Acknowledgments}
This work was supported in part by the National Natural Science Foundation of China (62206015, 62227801, U21B2048), the National Science and Technology Major Project (2022ZD0117901), and the Fundamental Research Funds for the Central Universities (FRF-TP-22-043A1).

\bibliography{aaai25}

\label{sec/6_supplement}
\newpage
\renewcommand\thesection{\Alph{section}}
\renewcommand\thefigure{\alph{figure}}
\renewcommand\thetable{\alph{table}}
\setcounter{section}{0}
\setcounter{figure}{0}
\setcounter{table}{0}

\section{State Space Model}

The state space model (SSM) is a type of linear time-invariant system that maps inputs to outputs through hidden layers. It is modeled by the following ordinary differential equations:
\begin{equation}
\begin{split}
    &h'(t)=Ah(t)+Bx(t),\\
    &y(t)=Ch(t).
\end{split}
\end{equation}

Where $A\in \mathbb{R}^{N\times N}$ is the evolution matrix, $B\in \mathbb{R}^{N\times 1}$ and $C\in \mathbb{R}^{1\times N}$ are the projection matrices. $N$ represents the sequence length. This continuous system is challenging to apply, while the SSM in Mamba serves as its discrete counterpart. It discretizes continuous parameters A and B into their discrete counterparts $\overline{A}$ and $\overline{B}$ using a time-scale parameter $\Delta$. This transformation typically employs the zero-order hold method:
\begin{equation}
\begin{split}
    &\overline{A}=exp(\Delta A),\\
    &\overline{B}=(\Delta A)^{-1}(exp(\Delta A)-I)\Delta B,\\
    &h_t=\overline{A}h_{t-1}+\overline{B}x_t,\\
    &y_t=Ch_t.
\end{split}
\end{equation}

The practical application of this discretization form is hindered by its inherent sequential nature. Nevertheless, it can be effectively represented through a convolution operation:
\begin{equation}
\begin{split}
    &\overline{K}=(C\overline{B},C\overline{AB},...,C\overline{A}^{N-1}\overline{B}),\\
    &y=x\ast \overline{K}.
\end{split}
\end{equation}

Where $\overline{K}\in \mathbb{R}^N$ denotes a structured convolution kernel, and $\ast$ indicates a convolution operation.

\section{Visualization}

\begin{figure*}[htbp]
  \centering
  \setlength{\abovecaptionskip}{0.15cm}
  \includegraphics[width=.8\linewidth]{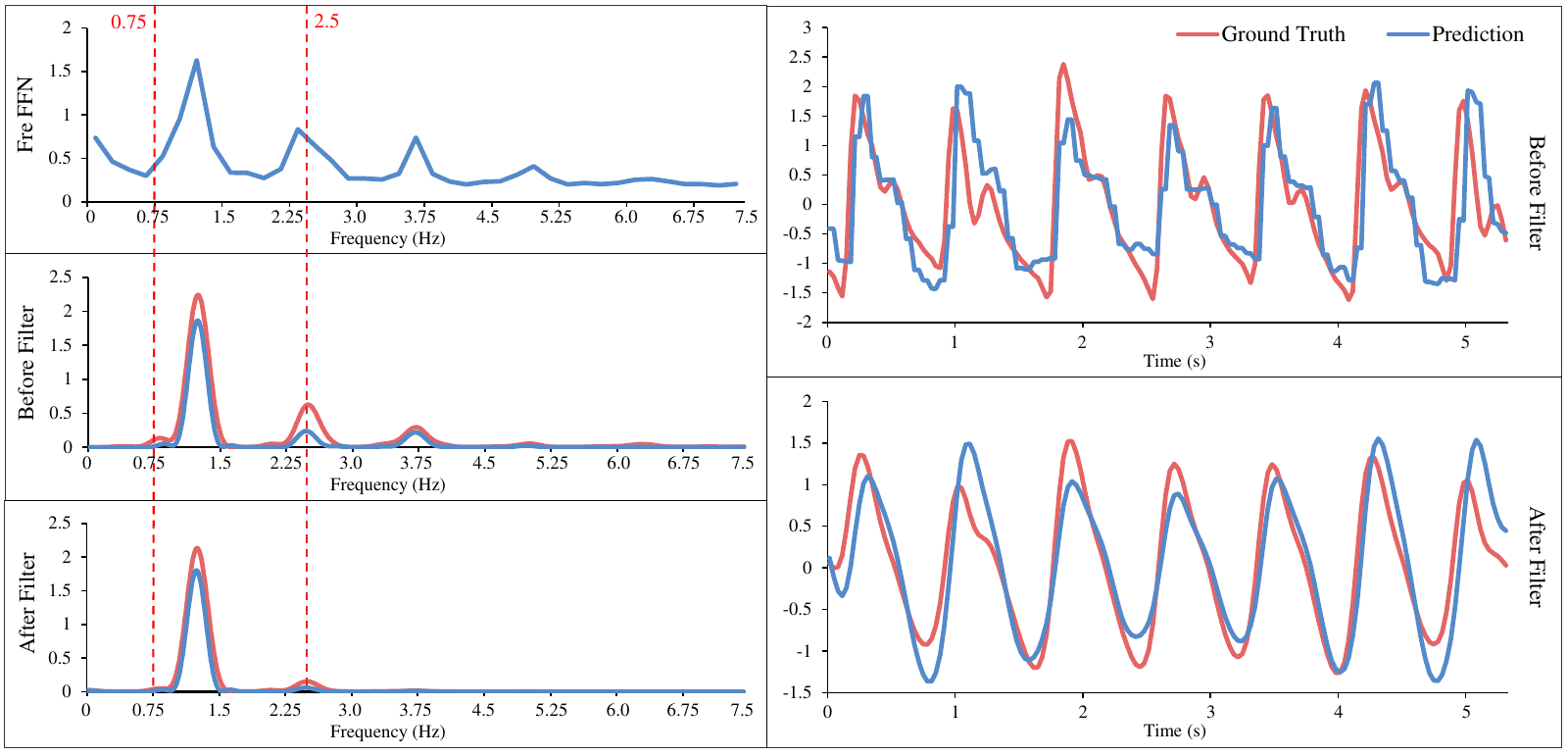}  
  \caption{An example of results on MMPD.}
  \label{fig:plot}
\end{figure*}

As shown in the left half of Figure \ref{fig:plot}, we visualize the spectrum of an example from MMPD. From top to bottom, these represent the averaging of spectra across all channels in the frequency domain feed-forward of the last block, the power spectral density of the PPG signal before bandpass filtering, and the power spectral density of the PPG after bandpass filtering. The range from 0.75 Hz to 2.5 Hz corresponds to the heart rate frequency band. Since the spectrum of the Fre FFN is obtained through FFT, the number of frequency points in the spectrum depends on the input length, resulting in a less smooth output. Nevertheless, it still accurately captures the frequency domain characteristics of the BVP ground truth.

As shown in the right half of Figure \ref{fig:plot}, an example of the PPG waveform is provided to demonstrate the efficacy and precision of our approach. Notably, our model robustly captures the peaks and variations of PPG signals, serving as the key basis for predicting heart rate from video data.
Additionally, the scatter plots and Bland-Altman plots in Figure \ref{fig:plot1} and Figure \ref{fig:plot2} further demonstrate the strong correlation between the predictions and the ground truth.
\vspace{1cm}

\section{Arbitrary Length Videos Input}

We attempt to test the trained RhythmMamba model on videos of arbitrary lengths, with the longest test video being limited to 60 seconds due to dataset constraints. As observed in Table \ref{tab:arbitrary}, the trained model enables seamless adaptation to video segments of any length without performance degradation. Only when the test length is reduced to 1 second (second row), which is close to or shorter than a single heartbeat, does some performance degradation occur. The results at various test lengths clearly demonstrate that RhythmMamba has effectively learned the periodic pattern of rPPG.

\begin{table}[htbp]
  % \small
  \centering
  \setlength{\abovecaptionskip}{0.25cm}
  \caption{Arbitrary length videos input.}
  \setlength{\tabcolsep}{4pt}
    \begin{tabular}{cccccc}
    \toprule
    Test Length & MAE$\downarrow$ & RMSE$\downarrow$ & MAPE$\downarrow$ & $\rho\uparrow$ & SNR$\uparrow$ \\
    \midrule
    160   & 3.16  & 7.27  & 3.37  & 0.84  & 4.74 \\
    30~(1s) & 5.50  & 11.65 & 6.02  & 0.66  & -1.70 \\
    80~(2.6s) & 3.11  & 6.61  & 3.30  & 0.87  & 3.09 \\
    300~(10s) & 3.21  & 7.43  & 3.42  & 0.83  & 6.11 \\
    600~(20s) & 3.10  & 7.61  & 3.32  & 0.82  & 8.15 \\
    900~(30s) & 3.51  & 7.30  & 3.70  & 0.84  & 8.56 \\
    1800~(60s) & 3.14  & 7.04  & 3.30  & 0.85  & 10.59 \\
    \bottomrule
    \end{tabular}%
  \label{tab:arbitrary}%
\end{table}%

\begin{figure*}[htbp]
  \centering
  \setlength{\abovecaptionskip}{0.15cm}
  \subfigure[Intra-dataset on MMPD.]{
  \includegraphics[width=.32\linewidth]{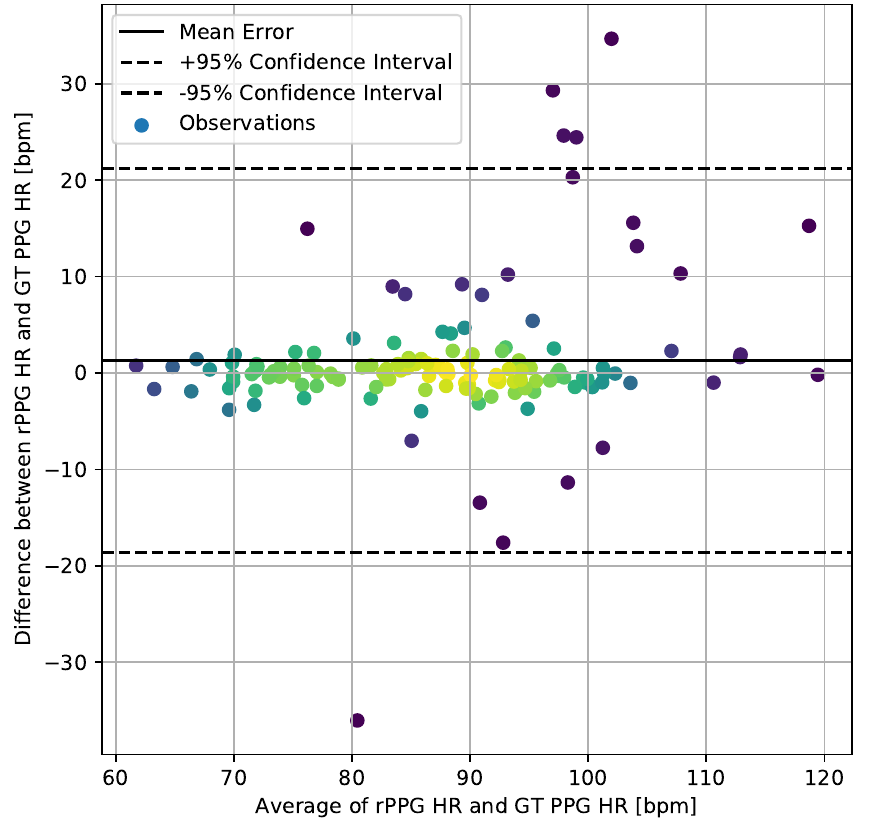}  
  }
  \subfigure[Cross-dataset on PURE-UBFC.]{
  \includegraphics[width=.315\linewidth]{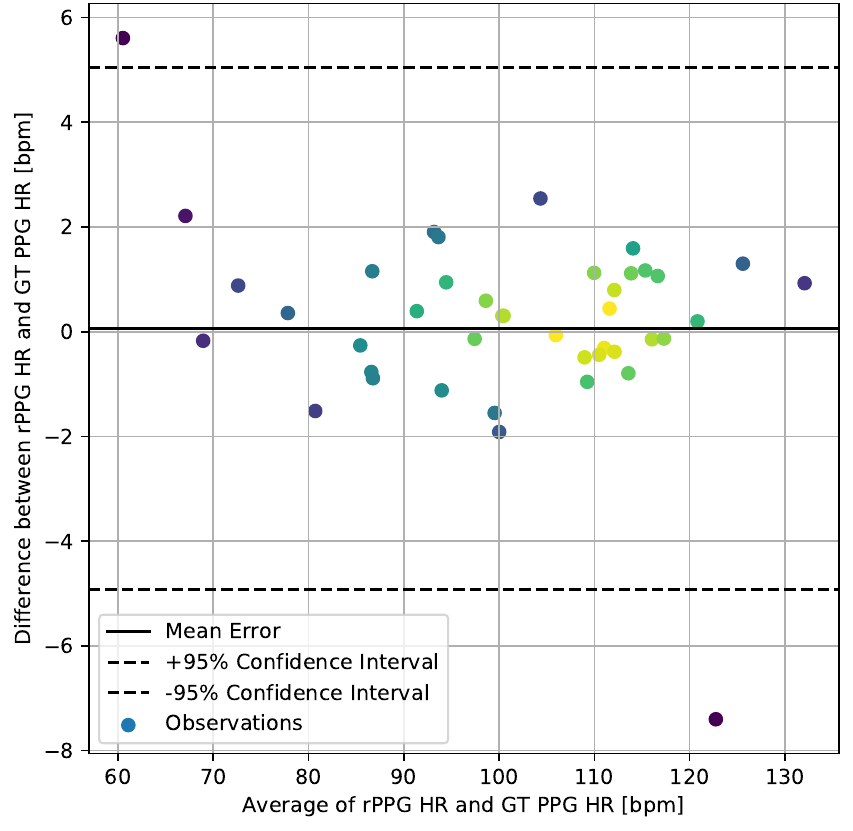}  
  }
  \subfigure[Cross-dataset on UBFC-PURE.]{
  \includegraphics[width=.32\linewidth]{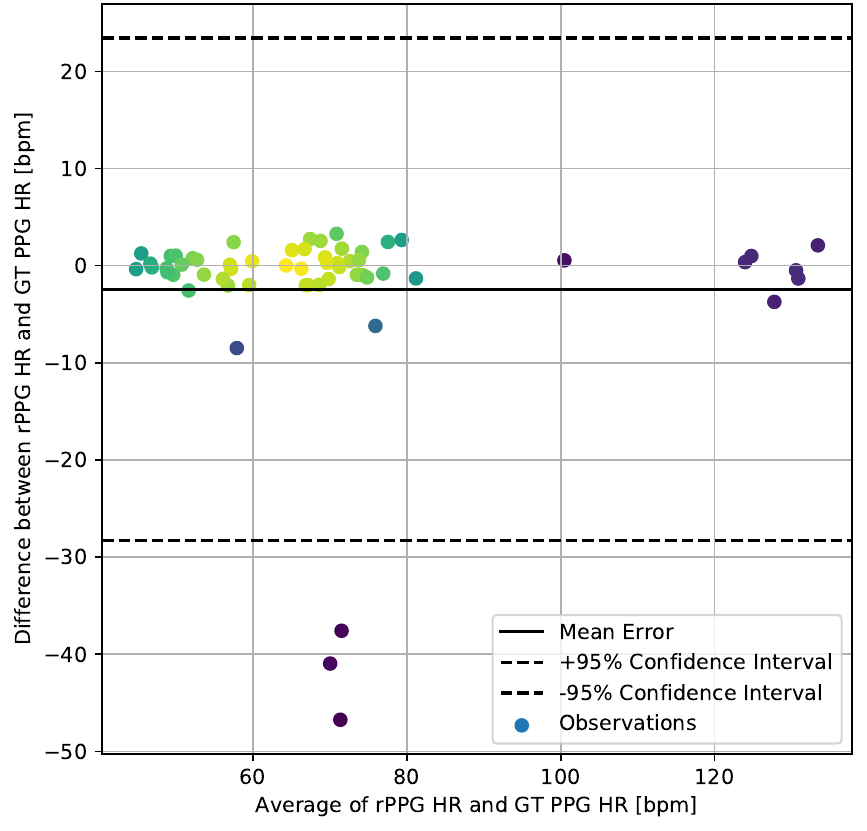}  
  }
  \caption{Bland-Altman plots of results.}
  \label{fig:plot1}
\end{figure*}

\begin{figure*}[htbp]
  \centering
  \setlength{\abovecaptionskip}{0.15cm}
  \subfigure[Intra-dataset on MMPD.]{
  \includegraphics[width=.32\linewidth]{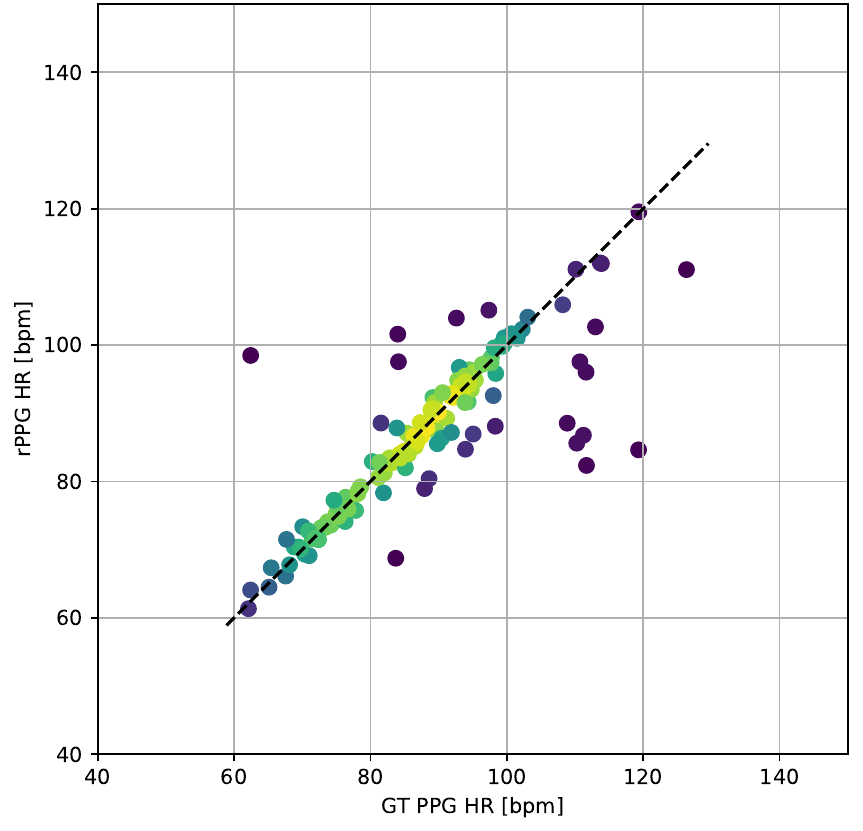}  
  }
  \subfigure[Cross-dataset on PURE-UBFC.]{
  \includegraphics[width=.32\linewidth]{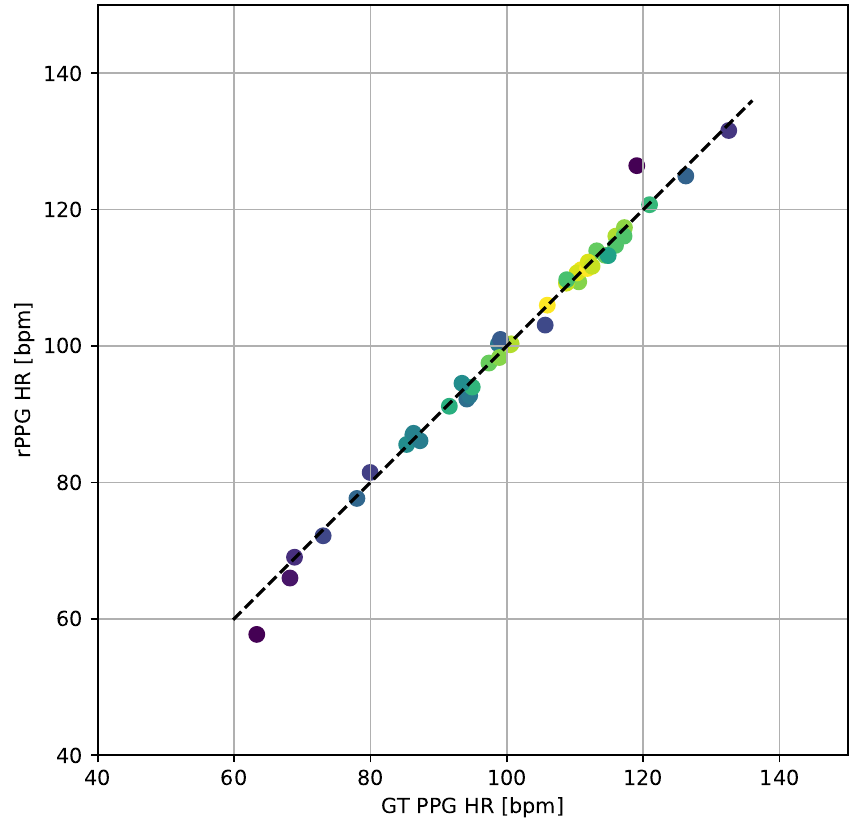}  
  }
  \subfigure[Cross-dataset on UBFC-PURE.]{
  \includegraphics[width=.32\linewidth]{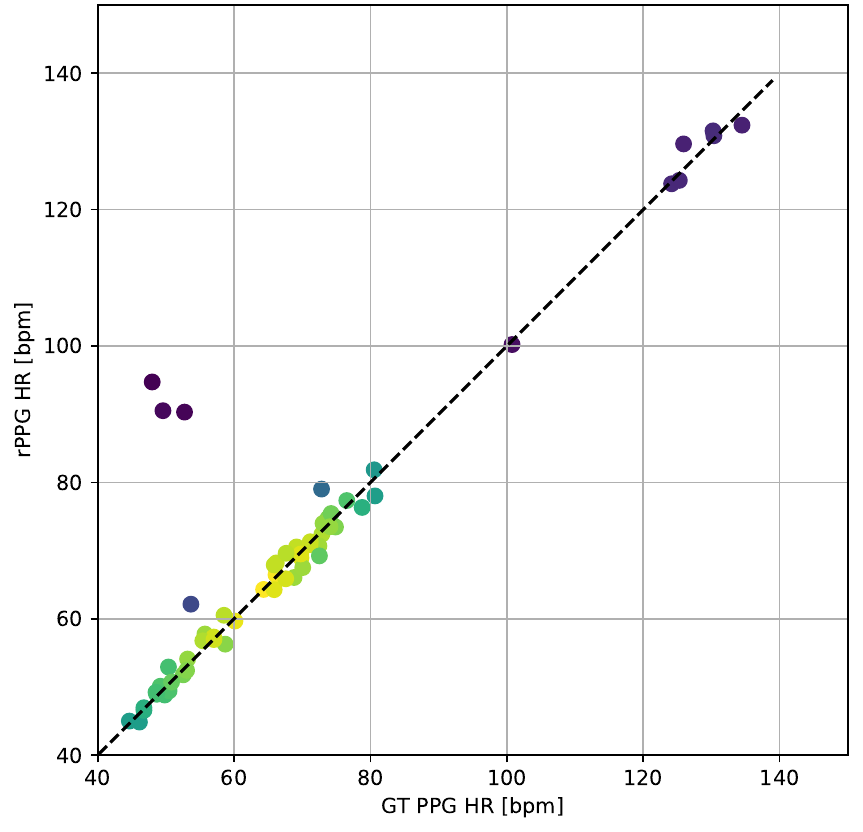}  
  }
  \caption{Scatter plots of results.}
  \label{fig:plot2}
\end{figure*}

\end{document}